\pdfoutput=1

\documentclass[11pt]{article}

\usepackage[]{EMNLP2023}

\usepackage{times}
\usepackage{latexsym}
\usepackage{multirow}
\usepackage{longtable}
\usepackage{float}

\usepackage[T1]{fontenc}

\usepackage[utf8]{inputenc}

\usepackage{microtype}

\usepackage{inconsolata}

\usepackage{graphicx}

\title{Long-form Question Answering: An Iterative Planning-Retrieval-Generation Approach}

\author{Pritom Saha Akash$^{1}$ $\quad$ Kashob Kumar Roy$^{1}$ $\quad$ Lucian Popa $^{2}$ $\quad$ Kevin Chen-Chuan Chang$^{1}$\\
$^{1}$University of Illinois at Urbana-Champaign, USA \\
$^2$IBM Research, USA \\
 \texttt{\{pakash2, kkroy2, kcchang\}@illinois.edu, lpopa@us.ibm.com}
}

\begin{document}
\maketitle

\begin{abstract}
Long-form question answering (LFQA) poses a challenge as it involves generating detailed answers in the form of paragraphs, which go beyond simple yes/no responses or short factual answers. While existing QA models excel in questions with concise answers, LFQA requires handling multiple topics and their intricate relationships, demanding comprehensive explanations. Previous attempts at LFQA focused on generating long-form answers by utilizing relevant contexts from a corpus, relying solely on the question itself. However, they overlooked the possibility that the question alone might not provide sufficient information to identify the relevant contexts. Additionally, generating detailed long-form answers often entails aggregating knowledge from diverse sources. To address these limitations, we propose an LFQA model with iterative Planning, Retrieval, and Generation. This iterative process continues until a complete answer is generated for the given question. From an extensive experiment on both an open domain and a technical domain QA dataset, we find that our model outperforms the state-of-the-art models on various textual and factual metrics for the LFQA task.
\end{abstract}
\section{Introduction}
\label{sec:introduction}
Question answering (QA) is a computational task that involves providing a relevant and accurate response to a question expressed in natural language. A considerable amount of progress has been made in open-domain question answering, specifically in settings where questions are answerable with short phrases and entities. For example, significant advancements have been made in factoid question-answering (QA) research, which has yielded impressive results with the creation of comprehensive datasets like SQuAD \cite{rajpurkar2018know} and MS MARCO \cite{nguyen2016ms}, as well as the utilization of transformer-based models such as ALBERT \cite{lan2019albert}. In many cases, these models have even demonstrated the ability to outperform human performance. However, while short-form question answering has proven to be effective for simple factual questions, it often falls short when it comes to complex and nuanced questions that require more comprehensive and detailed responses. 

One of the main challenges of LFQA is that there is not much data available for this task.  One prominent dataset used for this purpose is the ELI5 dataset \cite{fan2019eli5}. It comprises questions asked on the "Explain Like I'm Five" Reddit forum, along with corresponding answers in paragraph form. However, the questions in ELI5 tend to be broad (e.g., "How do animals see different colors?"), and multiple valid approaches can be taken to answer them. This multiplicity makes it difficult to establish objective criteria for assessing the quality of answers. In a study by \cite{krishna2021hurdles}, various hurdles in effectively leveraging this dataset for meaningful advancements in modeling were highlighted, including the absence of reliable evaluation metrics.

In addition to the limitation of available datasets, the existing models for LFQA fall short in performance. For example, the KILT benchmark, recently introduced by \cite{petroni2020kilt}, is a framework that evaluates retrieval-augmented models on various knowledge-intensive tasks, such as ELI5. It assesses LFQA models based on the quality of their generated answers (measured using ROUGE-L against reference answers) as well as the relevance of retrieved documents (measured using R-precision against human-annotated relevant documents). However, the utilization of retrieved contexts, such as passages or documents by models on the KILT leaderboard, is found to be minimal, according to the investigation conducted by \cite{krishna2021hurdles}. This lack of utilization poses a challenge for retrieval-augmented models aiming to enhance their performance in LFQA tasks. More specifically, the retrieved contexts fail to contribute significantly to the acquisition of new information necessary for generating comprehensive answers, impeding the models' progress.

One of the main reasons for the above problem is that previous models retrieve relevant contexts from a provided knowledge source or corpus, relying solely on the question itself. They overlooked the possibility that the question alone might not provide sufficient information to identify the relevant contexts within the corpus. Additionally, generating detailed long-form answers often entails aggregating knowledge snippets from diverse sources to provide a comprehensive explanation. To overcome these limitations, we propose a Long-Form Question Answering (LFQA) model with \textit{iterative planning, retrieve, and generation} (\textbf{IPRG}) approach. The idea behind our model is that even if the question does not have enough information for a complete answer generation, we may use that as an initial query to construct a plan, retrieve contexts, and generate a preliminary answer. This preliminary answer can then provide new hints for gathering further information by the next step of planning and retrieving. Consequently, it will generate more detailed answers for the LFQA task.

To evaluate the performance of the proposed models, we conduct both quantitative and qualitative evaluations on two datasets from two different domains. As specified before the limitation of the ELI5 dataset, we use two datasets with more definitive long-form answers from both an open domain (i.e., wikiHow) and a specific technical domain (i.e., apple exchange dataset).

\section{Methodology}
\label{sec:method}
\begin{figure*}[ht]
    \centering
    \includegraphics[height=0.4\textwidth, width=1.0\textwidth]{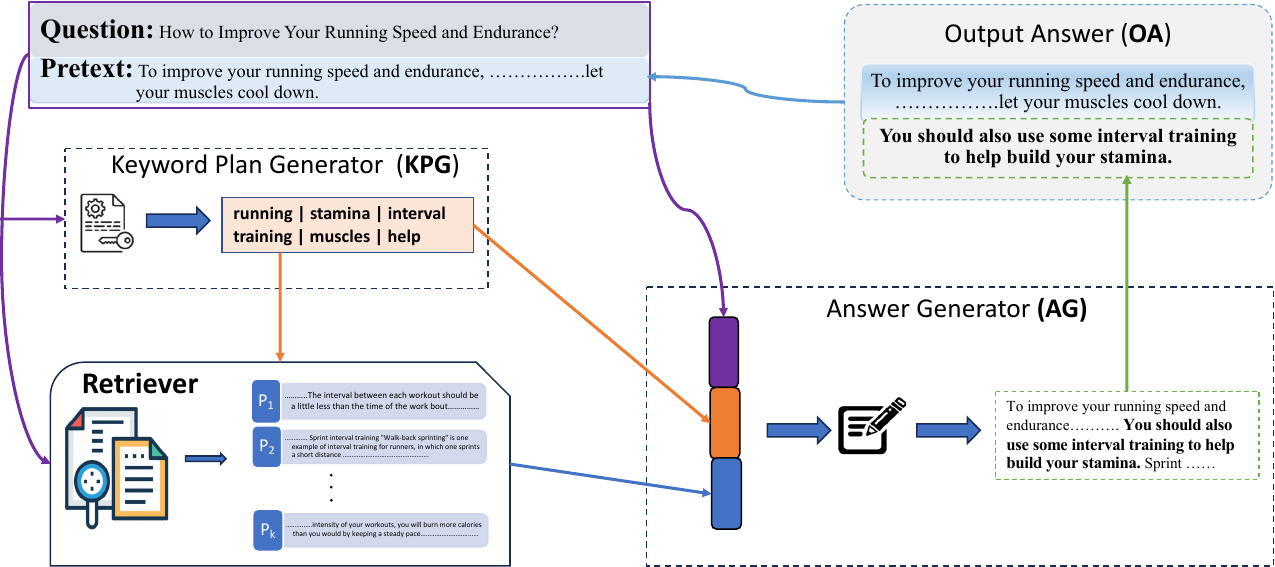}
    \caption{\textbf{IPRG}: At \textit{i}-th iteration, \textbf{KPG} produces a set of keywords from the question and pretext. Then, \textbf{Retriever} retrieves \textit{k} passages by using the question, pretext, and plan. Finally, the \textit{i}-th sentence in the answer paragraph generated by \textbf{AG} is appended to \textbf{OA}. The \textbf{OA} serves as pretext at the next iteration. }
    \label{fig:model_overview}
\end{figure*}

The proposed IPRG model addresses the task of LFQA where it takes as input a question $q$ and a corpus $\mathcal{C}$. The corpus can be obtained through methods such as web search or by using a static corpus like Wikipedia. The model consists of three modules: 1) A Keyword Plan Generator $p(w_i|[q;y_{1:i-1}])$ that generates a keyword plan $w_i$ consisting of a set of keywords for the next answer sentence given the concatenation of the question $q$ and already generated answer sentences $y_{1:i-1}$, 2) A Retriever $p(C_i|[q;w_i])$ that retrieves top $k$ passages as contexts $C_i$ from $\mathcal{C}$ for supporting the next answer sentence generation based on the question $q$ and current keyword plan $w_i$, and 3) an Answer Generator $p(y_i|[q;k_i;C_i])$ that generate next sentence for answer given the question, keyword plan and retrieved passages. In the following subsections, we will describe each of these modules, followed by how they are combined and trained for the LFQA task. The overview of the proposed architecture is shown in Figure \ref{fig:model_overview}. 

\subsection{Keyword Plan Generator}

A long-form answer demands detailed information. However, the conventional text generation models generally can generate short text with informativeness and coherence and tend to hallucinate or be repetitive if we force them to generate longer texts. One possible solution for this can be iteratively generating short text (i.e., sentence) at each iteration rather than generating the whole output at once. However, it still  need to solve the problem of generating repetitive sentences. To solve this issue, we predict some key points that will be used to guide the generation of the next answer sentence at each iteration. In other words, we plan to generate some future keywords that will be discussed in the next round of answer sentences. And these keywords also help find relevant contexts from the given corpus in the subsequent step. To do so, we frame this as a text-to-keyword generation (or prediction) problem. 

To generate an answer at each iteration, we first plan by finding what key points need to be discussed in the next sentence of the answer, and these key points also help find relevant contexts from the given corpus. We define this as a keyword generation task, where given a prompt, it generates a set of keywords. Initially, the prompt is the question itself. In the subsequent iterations, the generated answers portions get concatenated with the question to make the next prompt for generating the next set of keywords. More specifically, the Keyword Plan Generator $p(w_i|[q;y_{1:i-1}])$ that generates a keyword plan $w_i$ consisting of a set of keywords for the next answer sentence given the concatenation of the question $q$ and already generated answer sentences $y_{1:i-1}$.

To train the keyword plan generator, we convert each of the QA pairs in the training dataset into some texts to keyword-set pairs. For example, 
a question: “How to put or move downloaded files in different folders depending on file type?" and the first sentence of the ground truth answer is “I don't know of any Safari extension doing this but you may use Automator to create a folder action attached to your preferred download folder sorting files according to their extension or kind to various folders.” To convert the above pair into a training sample of the keyword plan generator, we first use an existing keyword extraction model to first extract important keywords from the first sentence of the ground truth answer, which is, for instance, Keys = ['preferred download folder', 'safari extension', 'various folders', 'folder action', 'use']. We use these extracted Keys as ground truth keywords for the Question. So the task in training is, given the Question, can we generate the Keys? In other words, starting with the Question itself, we generate the keywords of the first answer sentence using a seq2seq model. Subsequently, we add each sentence of the answer one after another as input to predict the keywords of the next answer sentence. We initialized with pretrained BART \cite{lewis2019bart} for the text-to-keyword generation task.

\subsection{Retriever}

At each step of the iterative process, given the question and generated keyword plan, we retrieve relevant contexts (e.g., passages or sentences) from the given corpus. Specifically, we concatenated the question, pretext, and generated keywords as a query to retrieve the top $k$ contexts from the corpus. We use an existing dense passage retrieval (DPR) \cite{karpukhin2020dense} model for this purpose.

\subsection{Answer Generator}
In this module, we ensemble the question, pretext, generated keywords, and retrieved contexts as input for the final answer generation model, which is another seq2seq model initialized using BART \cite{lewis2019bart}. Now, for the answer generation, one possible approach can be only generating the next sentence rather than the whole answer at a time. However, this may make the generated sentences disconnected from each other and which may result in the repetition of sentences in the generated answer. To solve this problem, we generate the paragraph length instead of a single sentence and append the first new sentence in the final output sentence. 
To train this seq2seq model, we use the answer sentences up to the next answer sentences as ground-truth answers. 

\section{Experiments}
\label{sec:experiments}

\textbf{WikiHowQA dataset: }
 We prepare a novel long-form question-answering dataset, \textbf{WikiHowQA}, based on the WikiHow knowledge base~\footnote{https://www.wikihow.com/Main-Page}. Each article title is a \textbf{“How to”} question. Field experts have written these articles and provided a coherent paragraph summary for each one. Unlike summarization datasets~\cite{cohen-etal-2021-wikisum, koupaee2018wikihow}, we design this dataset for open-domain long-form question answering tasks, where the paragraph summaries of articles are served as  the long-form answers to the title questions (as shown in Figure~\ref{fig:example}), and Wikipedia dump is used as knowledge corpus to retrieve relevant contexts. More details can be found in Appendix \ref{sec:wikihow_details}.

\textbf{Apple Exchange Dataset: }
To evaluate the proposed method in a technical domain, we use the Apple Exchange dataset adopted from a large COALA dataset~\cite{ruckle2019coala}. The answers to these technical questions require deeper technical knowledge compared to WikiHowQA. Thus Wikipedia dumps are not sufficiently informative to answer these questions. We retrieve the top 10 sentences relevant to each ground truth answer sentence in Google search, excluding sentences from the StackExchange website, and compile them into a knowledge corpus. This corpus can be easily updated online. For comparison purposes, we create a static corpus by crawling the web-search results.

\subsection{Comparison with Baselines}
\textbf{Baselines:} We compare our models with i) pretrained models such as GPT2-XL~\cite{radford2019language} and T5-3b~\cite{2020t5}, ii) Sequence-to-Sequence BART~\cite{lewis2019bart} model, retrieval-augmented generation DPR+BART~\cite{petroni2020kilt} model, Fusion in Decoder (FiD) model ~\cite{izacard2020leveraging} and MDR~\cite{xiong2021answering}. 
We fine-tune both BART, DPR+BART, FiD, and MDR on each target datasets and report results in Table~\ref{tab:results}.

\noindent \textbf{IRG}: A variant of IPRG by excluding Keyword Plan Generator in which both Retriever and Answer Generator do not take any sequence of keywords as input. 

\begin{table*}[]
\centering
\begin{tabular}{|c|cccc|cccc|}
\hline
\multirow{3}{*}{Models} & \multicolumn{4}{c|}{WikiHow QA}                                                          & \multicolumn{4}{c|}{Apple Exchange}                                                      \\ \cline{2-9} 
                        & \multicolumn{2}{c|}{Rouge-1}                          & \multicolumn{2}{c|}{Rouge-L}     & \multicolumn{2}{c|}{Rouge-1}                          & \multicolumn{2}{c|}{Rouge-L}     \\ \cline{2-9} 
                        & \multicolumn{1}{c|}{Recall} & \multicolumn{1}{c|}{F1} & \multicolumn{1}{c|}{Recall} & F1 & \multicolumn{1}{c|}{Recall} & \multicolumn{1}{c|}{F1} & \multicolumn{1}{l|}{Recall} & F1 \\ \hline
GPT2                    & \multicolumn{1}{c|}{4.66}       &  8.19  & \multicolumn{1}{|c|}{4.41}       &  7.75       & \multicolumn{1}{l|}{0.27}       & \multicolumn{1}{l|}{0.47}   & \multicolumn{1}{l|}{0.27}       &    0.47\\ 
T5                      & \multicolumn{1}{c|}{4.02}       & 6.68   & \multicolumn{1}{|c|}{3.78}       &  6.26       & \multicolumn{1}{l|}{0.30}       & \multicolumn{1}{l|}{0.49}   & \multicolumn{1}{l|}{0.30}       &   0.49 \\ \hline
BART                    & \multicolumn{1}{c|}{28.24}       &  32.49  & \multicolumn{1}{|c|}{26.40}       &  30.39   & \multicolumn{1}{c|}{11.81}       &  16.23  & \multicolumn{1}{|c|}{10.45}       &  14.36  \\ 
FiD              & \multicolumn{1}{c|}{25.19}       & 31.59   & \multicolumn{1}{|c|}{23.72}       &  29.75   & \multicolumn{1}{c|}{6.31}       & 9.48   & \multicolumn{1}{|c|}{5.79}       &  8.61 \\

DPR + BART              & \multicolumn{1}{c|}{28.71}       & 32.74   & \multicolumn{1}{|c|}{26.78}       &  30.54   & \multicolumn{1}{c|}{12.88}       & 17.22   & \multicolumn{1}{|c|}{11.32}       &  15.16 \\

MDR              & \multicolumn{1}{c|}{33.50}       & \underline{33.30}   & \multicolumn{1}{|c|}{31.01}       &  30.40   & \multicolumn{1}{c|}{22.74}       & 21.50   & \multicolumn{1}{|c|}{20.76}       &  19.56
\\ \hline
\textbf{IRG}  & \multicolumn{1}{c|}{\underline{33.68}}       &  3.29  & \multicolumn{1}{|c|}{\underline{31.30}}       &  \underline{30.99}   & \multicolumn{1}{c|}{\underline{23.88}}       &  \underline{21.78}  & \multicolumn{1}{|c|}{\underline{21.70}}       &  \underline{19.75} \\ 
\textbf{IPRG}               & \multicolumn{1}{c|}{\textbf{35.36}}       &  \textbf{33.65}  & \multicolumn{1}{|c|}{\textbf{32.88}}       &   \textbf{31.25}  & \multicolumn{1}{c|}{\textbf{24.73}}       &  \textbf{22.13}  & \multicolumn{1}{|c|}{\textbf{23.63}}       &   \textbf{20.4}2  \\ \hline
\end{tabular}
\caption{Comparison with baselines (\textbf{Best} and \underline{2$^{nd}$ best} score)}
\label{tab:results}
\end{table*}

\noindent\textbf{Results:}
Both IPRG and IRG consistently outperform all the baselines in all metrics.   Specifically, they outperform in recall scores (R-1 \& R-L) by a large margin. In LFQA, the answers demand a larger coverage of ground truth information. In other words, the higher recall values imply more detailed answers with accurate information. 


\textbf{IPRG generates more Entailed less Contradictory answer.}
Not only comparing performances by measuring the text overlapping with references, but we also compute entailment scores to evaluate how much the model's generated answers are logically aligned to the ground truth. We leverage a pretrained BART-large-mnli model (available on Huggingface\footnote{https://huggingface.co/facebook/bart-large-mnli}) where the generated answers are considered as hypothesis and ground-truth as the NLI premise. This model calculates the scores indicating the logical entailment, contradiction, or neutrality of a hypothesis with respect to the premise. As reported in Table~\ref{tab:entailment}, IPRG achieves more entailment scores as well as less contradictory scores in both datasets. Whereas, DPR+BART generated answers contain more contradiction. This happens because the  contexts retrieved using only questions might lack relevant important information. 
\begin{table}[h]
\setlength{\tabcolsep}{3pt} 
\centering
\resizebox{1.0\linewidth}{!}{%
\begin{tabular}[]{|c|cc|cc|}
\hline
\multirow{3}{*}{Models} & \multicolumn{2}{c|}{WikiHow QA}                                                                                   & \multicolumn{2}{c|}{Apple Exchange}                                                                               \\ \cline{2-5} 
                        & \multicolumn{2}{c|}{NLI}                                        & \multicolumn{2}{c|}{NLI}                                        \\ \cline{2-3} \cline{4-5}
                        & \multicolumn{1}{c|}{Entailed} & \multicolumn{1}{c|}{Contradict}                          & \multicolumn{1}{c|}{Entailed} & \multicolumn{1}{c|}{Contradict}                         \\ \hline
DPR + BART              & \multicolumn{1}{c|}{7.29}         & \multicolumn{1}{c|}{20.37}                                  & \multicolumn{1}{c|}{10.00
}         & \multicolumn{1}{c|}{20.49}                                      \\ 
MDR              & \multicolumn{1}{c|}{6.07}         & \multicolumn{1}{c|}{12.48}                                  & \multicolumn{1}{c|}{9.63
}         & \multicolumn{1}{c|}{20.77}                                      \\ 
\textbf{IRG}  & \multicolumn{1}{c|}{7.09}         & \multicolumn{1}{c|}{14.86}                                   & \multicolumn{1}{c|}{9.52}         & \multicolumn{1}{c|}{19.56}                                    \\ 
\textbf{IPRG}              & \multicolumn{1}{c|}{\textbf{8.34}}         & \multicolumn{1}{c|}{\textbf{10.98}}                                 & \multicolumn{1}{c|}{\textbf{12.74}}         & \multicolumn{1}{c|}{\textbf{13.88}}                                     \\ \hline
\end{tabular}}
\caption{Factual Consistency Comparison}
\label{tab:entailment}
\end{table}

\textbf{Planning enhances relevant Contexts Retrieval resulting in Comprehensive Answers.} 
DPR+BART performs per BART, indicating that question-only retrieved contexts in a single pass have minimal impacts. On the other hand, iterative retrieval refines the contexts resulting in the promising performance of the IRG model. Furthermore, the performance of the IPRG model is boosted by the two-fold advantages: i) keyword planning helps the retrieval module to identify and find out important information and the answer generator module to efficiently employ the current retrieved knowledge into answers at each iteration; ii) the understanding and knowledge about questions are getting enriched and refined by iterative characteristics of planning-retrieval-generation process. Thus, the performance improvement is reflected in both Rouge scores and entailment scores.  Please refer to Section ~\ref{sec:workflow} in appendix for a detailed walkthrough of our whole process.  

\begin{figure}[h]
    \centering
    \includegraphics[width = 1.0\linewidth]{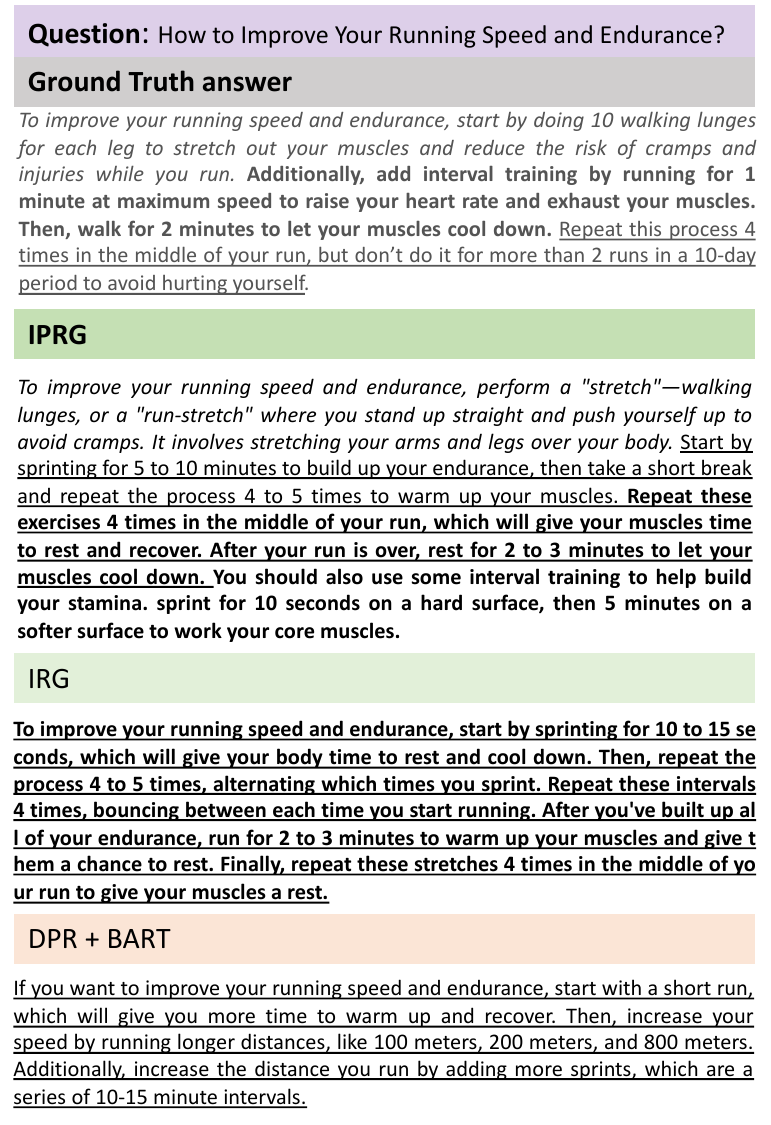}
    \caption{Comprehensiveness of answers w.r.t. three aspects: \underline{Strengthening}, \textbf{Interval} training, \textit{Stretching}.}
    \label{fig:example}
    \vspace{-5mm}
\end{figure}

\textbf{A Qualitative Case Study:} 
Our IPRG model can effectively capture relevant aspects to generate an elaborative answer for the target question. For instance, in Figure~\ref{fig:example}, the ground-truth answer mentions three ways to improve through: i) \underline{Strength training}, ii) \textbf{Interval training}, and iii) \textit{Stretching}. A well-detailed answer should include all aspects. We can see that both IPRG and IRG variants showed superior performance in capturing aspects. DPR+BART only includes details about \underline{strength} training but failed to capture the other two. On the other hand, IPRG perfectly captures all three aspects while \textit{stretching} aspect is absent in IRG's answer.

\subsection{Related Works}
\label{sec:related}
Open domain question answering is the task of answering questions by utilizing a knowledge base or corpus. In this setup,  passage retrieval is a key step to retrieving supporting documents before answering the question. A plethora of research has been conducted on retrieval-augmented question answering~\cite{lewis2020retrieval, petroni2020kilt, mao2021rider, izacard-grave-2021-leveraging, nguyen2016ms}. Existing works mostly focus on answering factoid questions. 

However, due to the recent emergent of generative models~\cite{radford2019language, lewis2019bart, 2020t5},  long-form question answering has become an active research area crowded by lots of unavoidable challenges such as factual hallucination, retrieving relevant context, and fluent and logically consistent answer generation. RAG~\cite{lewis2020retrieval}, an end-to-end retrieval-based generative model, which incorporates DPR~\cite{karpukhin2020dense} to retrieve supporting passages and then employ BART~\cite{lewis2019bart} to generate long-form answers from questions concatenated with evidence documents. Instead of concatenating document texts, FID~\cite{izacard-grave-2021-leveraging} encodes first retrieved documents independently  and then fuses representations into a decoder to generate answers. RBG~\cite{su-etal-2022-read} combines outputs from fusion-in-decoder module and machine reading comprehension module by following the mechanism of pointer-generator network~\cite{see2017get}. To improve the contexts, Re$^2$G~\cite{glass-etal-2022-re2g} augment RAG architecture by adding a learnable reranker module to select top k passages from a pool of passages from multiple retrievers. It is to be noted that all of these models perform their retrieval process once using questions as queries and generate answers in a single hop.

Recently, several multi-hop question-answering models have been introduced in~\cite{xiong2020answering, wang2022locate,yavuz2022modeling}, but are limited to factoid question answering. 
\section{Conclusion}
\label{sec:conclusion}

This paper introduces a new approach for long-form question-answering tasks by iteratively planning content through keywords, retrieving contexts from a corpus, and generating answers using all the available information. Unlike existing long-form question-answering models that suffer from underutilization of retrieved contexts, our model demonstrates better retrieval of relevant contexts from diverse sources feeding the generated answer with iterative refinement. Experiments on multiple datasets show the superiority of our model over other state-of-the-art models.  
\section{Limitation}
\label{sec:limitation}
Firstly, the proposed model uses a keyword planning module that depends on existing keyword extraction methods for training the module. Therefore, the error from the keyword extraction method may propagate to the other modules, resulting inconsistent and irrelevant answer generation. Developing a better task-dependent keyword planning generation that does not rely on the existing keyword extraction method can be a future research direction to improve the current model. Moreover, not all kinds of keywords are important as content plans for future answer generation. Therefore, automatically identifying informative keywords, in this case, can also improve the quality of the result.

Secondly, while training, the proposed IPRG model separately deals with each module which also makes it susceptible to error propagation. There is no scope to learn from one module's error to refine the model to another. Therefore, developing a joint end-end model can be another future research direction in this regard.

\bibliography{anthology}

\begin{thebibliography}{29}
\expandafter\ifx\csname natexlab\endcsname\relax\def\natexlab#1{#1}\fi

\bibitem[{Cohen et~al.(2021)Cohen, Kalinsky, Ziser, and
  Moschitti}]{cohen-etal-2021-wikisum}
Nachshon Cohen, Oren Kalinsky, Yftah Ziser, and Alessandro Moschitti. 2021.
\newblock \href {https://doi.org/10.18653/v1/2021.acl-short.28} {{W}iki{S}um:
  Coherent summarization dataset for efficient human-evaluation}.
\newblock In \emph{Proceedings of the 59th Annual Meeting of the Association
  for Computational Linguistics and the 11th International Joint Conference on
  Natural Language Processing (Volume 2: Short Papers)}, pages 212--219,
  Online. Association for Computational Linguistics.

\bibitem[{Coleman and Liau(1975)}]{coleman1975computer}
Meri Coleman and Ta~Lin Liau. 1975.
\newblock A computer readability formula designed for machine scoring.
\newblock \emph{Journal of Applied Psychology}, 60(2):283.

\bibitem[{Dale and Chall(1948)}]{dale1948formula}
Edgar Dale and Jeanne~S Chall. 1948.
\newblock A formula for predicting readability: Instructions.
\newblock \emph{Educational research bulletin}, pages 37--54.

\bibitem[{Fan et~al.(2019)Fan, Jernite, Perez, Grangier, Weston, and
  Auli}]{fan2019eli5}
Angela Fan, Yacine Jernite, Ethan Perez, David Grangier, Jason Weston, and
  Michael Auli. 2019.
\newblock Eli5: Long form question answering.
\newblock \emph{arXiv preprint arXiv:1907.09190}.

\bibitem[{Farr et~al.(1951)Farr, Jenkins, and
  Paterson}]{farr1951simplification}
James~N Farr, James~J Jenkins, and Donald~G Paterson. 1951.
\newblock Simplification of flesch reading ease formula.
\newblock \emph{Journal of applied psychology}, 35(5):333.

\bibitem[{Glass et~al.(2022)Glass, Rossiello, Chowdhury, Naik, Cai, and
  Gliozzo}]{glass-etal-2022-re2g}
Michael Glass, Gaetano Rossiello, Md~Faisal~Mahbub Chowdhury, Ankita Naik,
  Pengshan Cai, and Alfio Gliozzo. 2022.
\newblock \href {https://doi.org/10.18653/v1/2022.naacl-main.194} {{R}e2{G}:
  Retrieve, rerank, generate}.
\newblock In \emph{Proceedings of the 2022 Conference of the North American
  Chapter of the Association for Computational Linguistics: Human Language
  Technologies}, pages 2701--2715, Seattle, United States. Association for
  Computational Linguistics.

\bibitem[{Gunning et~al.(1952)}]{gunning1952technique}
Robert Gunning et~al. 1952.
\newblock Technique of clear writing.

\bibitem[{Izacard and Grave(2020)}]{izacard2020leveraging}
Gautier Izacard and Edouard Grave. 2020.
\newblock Leveraging passage retrieval with generative models for open domain
  question answering.
\newblock \emph{arXiv preprint arXiv:2007.01282}.

\bibitem[{Izacard and Grave(2021)}]{izacard-grave-2021-leveraging}
Gautier Izacard and Edouard Grave. 2021.
\newblock \href {https://doi.org/10.18653/v1/2021.eacl-main.74} {Leveraging
  passage retrieval with generative models for open domain question answering}.
\newblock In \emph{Proceedings of the 16th Conference of the European Chapter
  of the Association for Computational Linguistics: Main Volume}, pages
  874--880, Online. Association for Computational Linguistics.

\bibitem[{Karpukhin et~al.(2020)Karpukhin, O{\u{g}}uz, Min, Lewis, Wu, Edunov,
  Chen, and Yih}]{karpukhin2020dense}
Vladimir Karpukhin, Barlas O{\u{g}}uz, Sewon Min, Patrick Lewis, Ledell Wu,
  Sergey Edunov, Danqi Chen, and Wen-tau Yih. 2020.
\newblock Dense passage retrieval for open-domain question answering.
\newblock \emph{arXiv preprint arXiv:2004.04906}.

\bibitem[{Koupaee and Wang(2018)}]{koupaee2018wikihow}
Mahnaz Koupaee and William~Yang Wang. 2018.
\newblock Wikihow: A large scale text summarization dataset.
\newblock \emph{arXiv preprint arXiv:1810.09305}.

\bibitem[{Krishna et~al.(2021)Krishna, Roy, and Iyyer}]{krishna2021hurdles}
Kalpesh Krishna, Aurko Roy, and Mohit Iyyer. 2021.
\newblock Hurdles to progress in long-form question answering.
\newblock \emph{arXiv preprint arXiv:2103.06332}.

\bibitem[{Lan et~al.(2019)Lan, Chen, Goodman, Gimpel, Sharma, and
  Soricut}]{lan2019albert}
Zhenzhong Lan, Mingda Chen, Sebastian Goodman, Kevin Gimpel, Piyush Sharma, and
  Radu Soricut. 2019.
\newblock Albert: A lite bert for self-supervised learning of language
  representations.
\newblock \emph{arXiv preprint arXiv:1909.11942}.

\bibitem[{Lewis et~al.(2019)Lewis, Liu, Goyal, Ghazvininejad, Mohamed, Levy,
  Stoyanov, and Zettlemoyer}]{lewis2019bart}
Mike Lewis, Yinhan Liu, Naman Goyal, Marjan Ghazvininejad, Abdelrahman Mohamed,
  Omer Levy, Ves Stoyanov, and Luke Zettlemoyer. 2019.
\newblock Bart: Denoising sequence-to-sequence pre-training for natural
  language generation, translation, and comprehension.
\newblock \emph{arXiv preprint arXiv:1910.13461}.

\bibitem[{Lewis et~al.(2020)Lewis, Perez, Piktus, Petroni, Karpukhin, Goyal,
  K{\"u}ttler, Lewis, Yih, Rockt{\"a}schel et~al.}]{lewis2020retrieval}
Patrick Lewis, Ethan Perez, Aleksandra Piktus, Fabio Petroni, Vladimir
  Karpukhin, Naman Goyal, Heinrich K{\"u}ttler, Mike Lewis, Wen-tau Yih, Tim
  Rockt{\"a}schel, et~al. 2020.
\newblock Retrieval-augmented generation for knowledge-intensive nlp tasks.
\newblock \emph{Advances in Neural Information Processing Systems},
  33:9459--9474.

\bibitem[{Mao et~al.(2021)Mao, He, Liu, Shen, Gao, Han, and
  Chen}]{mao2021rider}
Yuning Mao, Pengcheng He, Xiaodong Liu, Yelong Shen, Jianfeng Gao, Jiawei Han,
  and Weizhu Chen. 2021.
\newblock Rider: Reader-guided passage reranking for open-domain question
  answering.
\newblock \emph{arXiv preprint arXiv:2101.00294}.

\bibitem[{Nguyen et~al.(2016)Nguyen, Rosenberg, Song, Gao, Tiwary, Majumder,
  and Deng}]{nguyen2016ms}
Tri Nguyen, Mir Rosenberg, Xia Song, Jianfeng Gao, Saurabh Tiwary, Rangan
  Majumder, and Li~Deng. 2016.
\newblock Ms marco: A human generated machine reading comprehension dataset.
\newblock \emph{choice}, 2640:660.

\bibitem[{Petroni et~al.(2020)Petroni, Piktus, Fan, Lewis, Yazdani, De~Cao,
  Thorne, Jernite, Karpukhin, Maillard et~al.}]{petroni2020kilt}
Fabio Petroni, Aleksandra Piktus, Angela Fan, Patrick Lewis, Majid Yazdani,
  Nicola De~Cao, James Thorne, Yacine Jernite, Vladimir Karpukhin, Jean
  Maillard, et~al. 2020.
\newblock Kilt: a benchmark for knowledge intensive language tasks.
\newblock \emph{arXiv preprint arXiv:2009.02252}.

\bibitem[{Radford et~al.(2019)Radford, Wu, Child, Luan, Amodei, Sutskever
  et~al.}]{radford2019language}
Alec Radford, Jeffrey Wu, Rewon Child, David Luan, Dario Amodei, Ilya
  Sutskever, et~al. 2019.
\newblock Language models are unsupervised multitask learners.
\newblock \emph{OpenAI blog}, 1(8):9.

\bibitem[{Raffel et~al.(2020)Raffel, Shazeer, Roberts, Lee, Narang, Matena,
  Zhou, Li, and Liu}]{2020t5}
Colin Raffel, Noam Shazeer, Adam Roberts, Katherine Lee, Sharan Narang, Michael
  Matena, Yanqi Zhou, Wei Li, and Peter~J. Liu. 2020.
\newblock \href {http://jmlr.org/papers/v21/20-074.html} {Exploring the limits
  of transfer learning with a unified text-to-text transformer}.
\newblock \emph{Journal of Machine Learning Research}, 21(140):1--67.

\bibitem[{Rajpurkar et~al.(2018)Rajpurkar, Jia, and Liang}]{rajpurkar2018know}
Pranav Rajpurkar, Robin Jia, and Percy Liang. 2018.
\newblock Know what you don't know: Unanswerable questions for squad.
\newblock \emph{arXiv preprint arXiv:1806.03822}.

\bibitem[{R{\"u}ckl{\'e} et~al.(2019)R{\"u}ckl{\'e}, Moosavi, and
  Gurevych}]{ruckle2019coala}
Andreas R{\"u}ckl{\'e}, Nafise~Sadat Moosavi, and Iryna Gurevych. 2019.
\newblock Coala: A neural coverage-based approach for long answer selection
  with small data.
\newblock In \emph{Proceedings of the AAAI Conference on Artificial
  Intelligence}, volume~33, pages 6932--6939.

\bibitem[{See et~al.(2017)See, Liu, and Manning}]{see2017get}
Abigail See, Peter~J Liu, and Christopher~D Manning. 2017.
\newblock Get to the point: Summarization with pointer-generator networks.
\newblock \emph{arXiv preprint arXiv:1704.04368}.

\bibitem[{Senter and Smith(1967)}]{senter1967automated}
RJ~Senter and Edgar~A Smith. 1967.
\newblock Automated readability index.
\newblock Technical report, Cincinnati Univ OH.

\bibitem[{Su et~al.(2022)Su, Li, Zhang, Shang, Jiang, Liu, and
  Fung}]{su-etal-2022-read}
Dan Su, Xiaoguang Li, Jindi Zhang, Lifeng Shang, Xin Jiang, Qun Liu, and
  Pascale Fung. 2022.
\newblock \href {https://doi.org/10.18653/v1/2022.findings-acl.61} {Read before
  generate! faithful long form question answering with machine reading}.
\newblock In \emph{Findings of the Association for Computational Linguistics:
  ACL 2022}, pages 744--756, Dublin, Ireland. Association for Computational
  Linguistics.

\bibitem[{Wang et~al.(2022)Wang, Wei, Fan, Zhang, and Huang}]{wang2022locate}
Siyuan Wang, Zhongyu Wei, Zhihao Fan, Qi~Zhang, and Xuanjing Huang. 2022.
\newblock Locate then ask: Interpretable stepwise reasoning for multi-hop
  question answering.
\newblock \emph{arXiv preprint arXiv:2208.10297}.

\bibitem[{Xiong et~al.(2021)Xiong, Li, Iyer, Du, Lewis, Wang, Mehdad, Yih,
  Riedel, Kiela, and Oguz}]{xiong2021answering}
Wenhan Xiong, Xiang Li, Srini Iyer, Jingfei Du, Patrick Lewis, William~Yang
  Wang, Yashar Mehdad, Scott Yih, Sebastian Riedel, Douwe Kiela, and Barlas
  Oguz. 2021.
\newblock \href {https://openreview.net/forum?id=EMHoBG0avc1} {Answering
  complex open-domain questions with multi-hop dense retrieval}.
\newblock In \emph{International Conference on Learning Representations}.

\bibitem[{Xiong et~al.(2020)Xiong, Li, Iyer, Du, Lewis, Wang, Mehdad, Yih,
  Riedel, Kiela et~al.}]{xiong2020answering}
Wenhan Xiong, Xiang~Lorraine Li, Srini Iyer, Jingfei Du, Patrick Lewis,
  William~Yang Wang, Yashar Mehdad, Wen-tau Yih, Sebastian Riedel, Douwe Kiela,
  et~al. 2020.
\newblock Answering complex open-domain questions with multi-hop dense
  retrieval.
\newblock \emph{arXiv preprint arXiv:2009.12756}.

\bibitem[{Yavuz et~al.(2022)Yavuz, Hashimoto, Zhou, Keskar, and
  Xiong}]{yavuz2022modeling}
Semih Yavuz, Kazuma Hashimoto, Yingbo Zhou, Nitish~Shirish Keskar, and Caiming
  Xiong. 2022.
\newblock Modeling multi-hop question answering as single sequence prediction.
\newblock \emph{arXiv preprint arXiv:2205.09226}.

\end{thebibliography}
\bibliographystyle{acl_natbib}

\appendix

\section{Appendix}
\label{sec:appendix}

\subsection{WikiHowQA dataset}
\label{sec:wikihow_details}
We collect article titles that have coherent paragraph summaries by filtering out those which have no paragraph-styled summaries or incomplete summaries. Consequently, the WikiHowQA dataset comprises 37,815 question-answer pairs. One of the key attributes of this dataset is that the answers are expert-written in plain English, which requires less world knowledge to understand and evaluate the answers. In order to measure the readability scores, we use classical readability metrics such as FKGL~\cite{farr1951simplification}, GFI~\cite{gunning1952technique}, ARI~\cite{senter1967automated}, CLI~\cite{coleman1975computer}, DCR~\cite{dale1948formula}. All metrics' generated scores indicate the number of years of formal education required for a native English speaker to understand the answer text. As reported in Table~\ref{tab:readability}, the scores for WikiHowQA answers are significantly smaller than that of the widely used ELI5 dataset.

 \begin{table}[h]
 \centering
 \small
\setlength{\tabcolsep}{3pt} 
\begin{tabular}{|c|ccccc|}
\hline
\multirow{2}{*}{Datasets} & \multicolumn{5}{c|}{Readability}                                                                                       \\ \cline{2-6} 
                          & \multicolumn{1}{c|}{FKGL} & \multicolumn{1}{c|}{DCR}  & \multicolumn{1}{c|}{ARI}   & \multicolumn{1}{c|}{CLI}  & GFI   \\ \hline
\textbf{WikiHowQA}                 & \multicolumn{1}{c|}{\textbf{8.70}} & \multicolumn{1}{c|}{\textbf{7.78}} & \multicolumn{1}{c|}{\textbf{9.12}}  & \multicolumn{1}{c|}{\textbf{7.58}} & \textbf{11.37} \\
ELI5                      & \multicolumn{1}{c|}{9.81} & \multicolumn{1}{c|}{8.38} & \multicolumn{1}{c|}{10.22} & \multicolumn{1}{c|}{8.97} & 12.70 \\ \hline
\end{tabular}
\caption{Readability scores for reference answers. Smaller value is more Readable.}
\label{tab:readability}
\end{table}

\begin{table*}[h]
\scriptsize
\begin{tabular}{|p{0.35\textwidth} | p{0.6\textwidth}|}
\hline
\textbf{Query}                                                                                                                    & \textbf{Retrieved top two Passages}                                 \\ \hline
\begin{tabular}{p{0.34\textwidth}}\textbf{Question}: How to Improve Your Running Speed and Endurance?\\ \\
\textbf{Pretext}: ""\\ \\
\textbf{Keywords}: walking lunges, run, endurance, stretch, cramps
\end{tabular} 
& \begin{tabular}{p{0.59\textwidth}}\textbf{P1}: including 2–3 days of core strengthening.  In the base phase the athlete now works on building cardiovascular \textbf{endurance} by having several long \textbf{runs} staying in heart rate zone 1-2 every week and each week adding slightly more mileage (using 10\% rule for safely increasing the mileage).  Core strengthening is continued in the base period. Once the base phase is complete and the athlete has sufficient \textbf{endurance}, the build period is needed to give the athlete the ability to hold a faster pace for the race duration. The build phase is where duration of runs is traded for intensity.  \\  \\

\textbf{P2}: \textbf{Stretching} does not appear to reduce the risk of injury during exercises, except perhaps a dynamic warm-up for runners. While running places extreme stress loads on the joints, static \textbf{stretching} can help to improve joint flexibility. However, this has not been proven to reduce risk of injury in the runners. A dynamic (stretching) warm up has been shown to help overall running performance.
One should avoid overtraining to prevent \textbf{cramps}. The calf muscles slowly respond to
\end{tabular} \\ \hline

\begin{tabular}{p{0.34\textwidth}}\textbf{Question}: How to Improve Your Running Speed and Endurance?\\ \\
\textbf{Pretext}: To improve your \textbf{running} speed and \textbf{endurance}, perform a "\textbf{stretch}"—\textbf{walking lunges}, or a "run-stretch" where you stand up straight and push yourself up to avoid \textbf{cramps}.\\ \\
\textbf{Keywords}: stretch, arms, muscles
\end{tabular} 
& \begin{tabular}{p{0.59\textwidth}}\textbf{P1}: leg muscles. Unlike the previous exercises, these are outdoor activities.
The calf exercises should be performed every 3–4 days. They can be alternated and interchanged to avoid getting used to the load. General Workout Tips Before any serious activity, including doing calf-building exercises, the muscles and joints should be properly warmed up.
Aerobic exercises should be performed at the beginning of a workout. A workout should end with calf strengthening exercises to stimulate their growth, and stretching that involves taking a wide step backward, placing the heel on the floor, and bending the torso forward. Each leg should be stretched for 10–20.  \\  \\

\textbf{P2}: must return to, and pause in, the correct starting position before continuing. If you rest on the ground or raise either hand or foot from the ground, your performance will be terminated. You may reposition your hands and/or feet during the event as long as they remain in contact with the ground at all times. Correct performance is important. You will have two minutes in which to do as many push-ups as you can." Sit-up "The sit-up event measures the endurance of the abdominal and hip-flexor muscles. On the command 'get set,' assume the starting position by lying on your.
\end{tabular} \\ \hline

\begin{tabular}{p{0.34\textwidth}}\textbf{Question}: How to Improve Your Running Speed and Endurance?\\ \\
\textbf{Pretext}: To improve your running speed and endurance, perform a "stretch"—walking lunges, or a "run-stretch" where you stand up straight and push yourself up to avoid cramps. It involves \textbf{stretching} your \textbf{arms} and legs over your body.\\ \\
\textbf{Keywords}: sprinting, repeat, process, minutes, times
\end{tabular} 
& \begin{tabular}{p{0.59\textwidth}}\textbf{P1}: in losing weight, staying in shape and improving body composition. Research suggests that the person of average weight will burn approximately 100 calories per mile run. Running increases one's metabolism, even after running; one will continue to burn an increased level of calories for a short time after the run. Different speeds and distances are appropriate for different individual health and fitness levels. For new runners, it takes time to get into shape. The key is consistency and a slow increase in speed and distance. While running, it is best to pay attention to how one's body feels.  \\  \\

\textbf{P2}: Many training programs last a minimum of five or six months, with a gradual increase in the distance run and finally, for recovery, a period of tapering in the one to three weeks preceding the race. For beginners wishing to merely finish a marathon, a minimum of four months of running four days a week is recommended. Many trainers recommend a weekly increase in mileage of no more than 10\%.
\end{tabular} \\ \hline

\begin{tabular}{p{0.34\textwidth}}\textbf{Question}: How to Improve Your Running Speed and Endurance?\\ \\
\textbf{Pretext}: To improve your running speed and endurance, perform a "stretch"—walking lunges, or a "run-stretch" where you stand up straight and push yourself up to avoid cramps. It involves stretching your arms and legs over your body. Start by \textbf{sprinting} for 5 to 10 \textbf{minutes} to build up your endurance, then take a short break and \textbf{repeat} the \textbf{process} 4 to 5 \textbf{times} to warm up your muscles. \\ \\
\textbf{Keywords}: run, repeat, muscles, reps, recover
\end{tabular} 
& \begin{tabular}{p{0.59\textwidth}}\textbf{P1}: a base for more intense workouts by strengthening the heart and increasing the muscles' ability to use oxygen, and to recover between hard workouts. Daniels recommends that most training miles are performed in E pace. Typical E runs include continuous runs up to about an hour. Marathon (M) pace At 80-85\% $HR_{max}$, this intensity is primarily aimed towards runners training for the marathon. The pace is one at which the runner hopes to compete. The pace can be included in other programs for a more intense workout, especially if the runner feels fresh and there is enough time to recover afterwards.
  \\  \\

\textbf{P2}: in losing weight, staying in shape and improving body composition. Research suggests that the person of average weight will burn approximately 100 calories per mile run. Running increases one's metabolism, even after running; one will continue to burn an increased level of calories for a short time after the run. Different speeds and distances are appropriate for different individual health and fitness levels. For new runners, it takes time to get into shape. The key is consistency and a slow increase in speed and distance. While running, it is best to pay attention to how one's body feels.

\end{tabular} \\ \hline

\begin{tabular}{p{0.34\textwidth}}\textbf{Question}: How to Improve Your Running Speed and Endurance?\\ 
\textbf{Pretext}: To improve your running speed and endurance, perform a "stretch"—walking lunges, or a "run-stretch" where you stand up straight and push yourself up to avoid cramps. It involves stretching your arms and legs over your body. Start by sprinting for 5 to 10 minutes to build up your endurance, then take a short break and repeat the process 4 to 5 times to warm up your muscles. \textbf{Repeat} these exercises 4 times in the middle of your \textbf{run}, which will give your \textbf{muscles} time to rest and \textbf{recover}.\\ \\  
\textbf{Keywords}: stretch, arms,  legs, run, repeat, times
\end{tabular} 
& \begin{tabular}{p{0.59\textwidth}}\textbf{P1}: must return to, and pause in, the correct starting position before continuing. If you rest on the ground or raise either hand or foot from the ground, your performance will be terminated. You may reposition your hands and/or feet during the event as long as they remain in contact with the ground at all times. Correct performance is important. You will have two minutes in which to do as many push-ups as you can." Sit-up "The sit-up event measures the endurance of the abdominal and hip-flexor muscles. On the command 'get set,' assume the starting position by lying on your  \\  \\

\textbf{P2}: in losing weight, staying in shape and improving body composition. Research suggests that the person of average weight will burn approximately 100 calories per mile run. Running increases one's metabolism, even after running; one will continue to burn an increased level of calories for a short time after the run. Different speeds and distances are appropriate for different individual health and fitness levels. For new runners, it takes time to get into shape. The key is consistency and a slow increase in speed and distance. While running, it is best to pay attention to how one's body feels.

\end{tabular} \\ \hline

\end{tabular}
\caption{Walk through IPRG model workflow with an example}
\label{tab:example_flow_1}
\end{table*}

\begin{table*}[h]
\scriptsize
\begin{tabular}{|p{0.35\textwidth} | p{0.6\textwidth}|}
\hline
\textbf{Query}                                                                                                                    & \textbf{Retrieved top two Passages}                                 \\ \hline

\begin{tabular}{p{0.34\textwidth}}\textbf{Question}: How to Improve Your Running Speed and Endurance?\\ 
\textbf{Pretext}: To improve your running speed and endurance, perform a "stretch"—walking lunges, or a "run-stretch" where you stand up straight and push yourself up to avoid cramps. It involves stretching your arms and legs over your body. Start by sprinting for 5 to 10 minutes to build up your endurance, then take a short break and repeat the process 4 to 5 times to warm up your muscles. Repeat these exercises 4 times in the middle of your run, which will give your muscles time to rest and recover. After your run is over, run for 2 to 3 minutes to let your muscles cool down.\\ 
\textbf{Keywords}: running, stamina, interval training, muscles, help
\end{tabular} 
& \begin{tabular}{p{0.59\textwidth}}\textbf{P1}: cope with the intensity, and to train for longer periods of time, this training is performed as interval training, hence the name. The interval between each workout should be a little less than the time of the work bout. Optimum intervals are 3–5 minutes long. There is no benefit to exceeding 5 minutes at this pace, under Daniels' theory, which means that despite the popularity of mile-repeats in many running groups, Daniels discourages them for people whose pace is slower than about 5:00/mile, preferring shorter intervals such as 1200 meters.  \\  \\

\textbf{P2}: (a minimum of 45 minutes). The development of aerobic and anaerobic capacities, and the adaptability of fartlek - to mimic running during specific sports - are characteristics it shares with other types of interval training. Sprint interval training "Walk-back sprinting" is one example of interval training for runners, in which one sprints a short distance (anywhere from 100 to 800 metres), then walks back to the starting point (the recovery period), to repeat the sprint a certain number of times. To add challenge to the workout, each of these sprints may start at predetermined time intervals - e.g. 200 metre.
One should avoid overtraining to prevent cramps. The calf muscles slowly respond to

\end{tabular} \\ \hline
\begin{tabular}{p{0.34\textwidth}}\textbf{Question}: How to Improve Your Running Speed and Endurance?\\ \\
\textbf{Pretext}: To improve your running speed and endurance, perform a "stretch"—walking lunges, or a "run-stretch" where you stand up straight and push yourself up to avoid cramps. It involves stretching your arms and legs over your body. Start by sprinting for 5 to 10 minutes to build up your endurance, then take a short break and repeat the process 4 to 5 times to warm up your muscles. Repeat these exercises 4 times in the middle of your run, which will give your muscles time to rest and recover. After your run is over, run for 2 to 3 minutes to let your muscles cool down. You should also use some \textbf{interval} \textbf{training} to help build your \textbf{stamina}. \\ \\
\textbf{Keywords}: run, sprint, muscles, time
\end{tabular} 
& \begin{tabular}{p{0.59\textwidth}}\textbf{P1}: workouts the day after interval sessions. Finally the race phase of the periodization approach is where the duration of the workouts decreases but intense workouts remain so as to keep the high lactate threshold that was gained in the build phase. In Ironman training, the race phase is where a long "taper" occurs of up to 4 weeks for highly trained Ironman racers.  A final phase is designated transition and is a period of time, where the body is allowed to recover from the hard race effort and some maintenance endurance training is performed so the high fitness level.  \\  \\

\textbf{P2}:(a minimum of 45 minutes). The development of aerobic and anaerobic capacities, and the adaptability of fartlek - to mimic running during specific sports - are characteristics it shares with other types of interval training. Sprint interval training "Walk-back sprinting" is one example of interval training for runners, in which one sprints a short distance (anywhere from 100 to 800 metres), then walks back to the starting point (the recovery period), to repeat the sprint a certain number of times. To add challenge to the workout, each of these sprints may start at predetermined time intervals - e.g. 200 metre.

\end{tabular} \\ \hline

\begin{tabular}{p{0.34\textwidth}}\textbf{Question}: How to Improve Your Running Speed and Endurance?\\ \\
\textbf{Answer}: To improve your running speed and endurance, perform a "stretch"—walking lunges, or a "run-stretch" where you stand up straight and push yourself up to avoid cramps. It involves stretching your arms and legs over your body. Start by sprinting for 5 to 10 minutes to build up your endurance, then take a short break and repeat the process 4 to 5 times to warm up your muscles. Repeat these exercises 4 times in the middle of your run, which will give your muscles time to rest and recover. After your run is over, run for 2 to 3 minutes to let your muscles cool down. You should also use some interval training to help build your stamina. \textbf{Sprint} for 10 seconds on a hard surface, then 5 minutes on a softer surface to work your core \textbf{muscles}. \\ \\
\end{tabular} 
& \begin{tabular}{p{0.59\textwidth}} \\  \\

\end{tabular} \\ \hline
\end{tabular}
\caption{Walk through IPRG model workflow with an example (as the continuation of Table~\ref{tab:example_flow_1})}
\label{tab:example_flow_2}
\end{table*}
\subsection{Model Workflow with an Example}
\label{sec:workflow}
In this section, we illustrate our model workflow using an example step-by-step iteratively shown in Figure~\ref{tab:example_flow_1} and~\ref{tab:example_flow_2}. Each row of these tables contains  the question, current pretext, and generated keywords at each iteration step in the first column. For convenience, we display the top 2 retrieved passages in the second column.

\end{document}